\documentstyle[colap]{article}

\makeatletter
 \def\submitted#1{\setbox\@tempboxa\vbox{\normalsize \tt \raggedright
    #1 \\ \hbox{}}
    \vspace{-2.5 cm} \usebox\@tempboxa \\
    \vspace{-\ht\@tempboxa} \vspace{2.5 cm}}
\makeatother

\title{\submitted{Appears in "Proceedings COLING90", Helsinki, August 1990}Centering theory and the  Italian pronominal system}
\author{
Barbara Di Eugenio \thanks{This research was supported by DARPA grant no. N0014-85-K0018.}\\
Department of Computer and Information Science\\
University of Pennsylvania\\
Philadelphia, PA\\
dieugeni@linc.cis.upenn.edu}
\date{}

\begin{document}

\bibliographystyle{alpha}

\pagestyle{empty}
\newtheorem{exa}{Ex.}

\maketitle
\begin{abstract}
In this paper, I give an account, in terms of centering theory
\cite{gjw86}, of some phenomena of pronominalization in Italian, in
particular the use of the null or the overt pronoun in subject
position.  After a general introduction to the Italian pronominal
system, I will review centering, and then show how the original rules
given in \cite{gjw86} have to be extended or modified.  Finally, I
will show that centering does not account for two phenomena: first,
the functional role of an utterance may override the predictions of
centering; second, a null subject can be used to refer to a whole
discourse segment. This latter phenomenon should ideally be explained
in the same terms that the other phenomena involving null subject are.
\end{abstract}

\section{The Italian pronominal system}

In Italian, there are two pronominal systems, characterized by a
different syntactic distribution: weak pronouns, that must always be
cliticized to the verb (e.g. {\sf la, lo, li, le} - respectively her,
accusative; him, accusative; them, masculine, accusative; them,
feminine, accusative or her, dative), and strong pronouns ({\sf lui,
lei, loro} - respectively he or him; she or her; they or them).  The
null subject can be considered as belonging to the system of weak
pronouns.  Notice that in Italian there is no neuter gender: nouns
referring to inanimate objects are masculine or feminine.  The weak
pronouns used in this case are those of the corresponding gender,
while, when a strong pronoun has to be used, paraphrase or deictics
are preferred.  A strong pronoun for inanimate objects does exist -
{\sf esso} for masculine, {\sf essa} for feminine, but it is not much
used in current Italian.

Weak and strong pronouns are often in complementary distribution,
as the following example shows - the contrast is between the use of
the null  or overt pronoun in subject position
\footnote{$\phi$ indicates a null subject and can be translated as an unstressed pronoun in English.
In all the examples I will be using, if a proper name ends in {\em -o}
or {\em -i}, it has a male referent; if it ends in {\em -a}, a female
referent. The translations I provide are literal and generally
word by word.}:

\boldmath
\footnotesize
\begin{exa}
\normalsize
\footnotesize
\begin{tabbing}
\hspace{1cm} 
\= a) \=  Quando {\bf Carlo}$_i$ ha incontrato {\bf Mario}$_j$, \\
\> \> When {\bf Carlo}$_i$ has  met {\bf Mario}$_j$,\\
\> \> $\phi_{i/*j}$  non {\bf gli}$_{*i/j}$
ha nemmeno detto ``ciao''.\\
\> \>  {\bf he}$_{i/*j}$ not {\bf to-him}$_{i/*j}$ has even said ``hi''.\\
\> b) Quando  {\bf Carlo}$_i$ ha incontrato {\bf Mario}$_j$, \\
\> \>  When {\bf Carlo}$_i$ has met {\bf Mario}$_j$, \\
\> \> {\bf lui}$_{*i/j}$  non {\bf gli}$_{i/*j}$
ha nemmeno detto ``ciao''.\\ \label{exa:carlo}
\> \> {\bf he}$_{*i/j}$ not {\bf to-him}$_{i/*j}$ has even said ``hi''.
\end{tabbing}
\end{exa}
\unboldmath
\normalsize

Notice the difference between sentences {\sf a} and {\sf b}: 
in {\sf a} the null pronoun in subject
position  refers to {\sf Carlo} and therefore {\sf gli} has to refer to 
{\sf Mario}; in {\sf b} reference is switched.
The overt pronoun {\sf lui} in subject position
requires its referent to be {\sf Mario}, and therefore {\sf gli} has to refer
to {\sf Carlo}.
 
There are some syntactic accounts of coreference phenomena in Italian, 
for example Calabrese's \cite{cal}. 
He starts from the
observation that weak pronouns are used in all those contexts
in which there is an expected referent for the pronoun itself; he claims
that we cannot use strong pronouns
in place of weak ones, and vice versa \footnote{Actually
Calabrese classifies pronouns as
{\em unstressed / stressed},
and not as {\em weak / strong}, 
but I think his terminology may lead the reader
to a wrong conclusion.
In fact, while the ``unstressed'' pronouns can never be stressed, the 
``stressed'' pronouns can, but not necessarily are.}.

To formalize the concept of {\em expected referent}, he resorts to the
notion of Thema, defined as {\em the subject of a primary
predication}, where {\em x is a primary predicate of y iff x and y
form a constituent which is either $\theta-$marked or [$+$INFL]}. \\
He then says that a pronoun in position of Thema is expected to have
another Thema as antecedent, and that if this coindexing occurs, the
pronoun must be a weak one.

Through these definitions and rules he manages to account for a wide
range of data, as far as single sentences are concerned, but when he
tries to extend them to discourse, their usefulness and predictive
power is not sufficient, and sometimes they give the wrong prediction.  
This is partly due to his very simplistic view of
discourse, which he considers as a conjunction of sentences.  Even for
those sentences in which this view is sufficient, the argument that
coreference depends only on the syntactic structure of the discourse
and that we cannot use a weak pronoun when
the theory predicts that a strong one is expected does not hold.
Consider the following example:

\boldmath
\footnotesize
\begin{exa}
\normalsize
\footnotesize
\begin{tabbing}
\hspace{1cm} 
\= D1) \= a) \= Ieri {\bf Carlo}$_i$ ha incontrato {\bf Mario}$_j$. \\
\> \> \> Yesterday {\bf Carlo}$_i$ has met {\bf Mario}$_j$.\\
\> \> b) $\phi_{i/*j}$  Non {\bf gli}$_{*i/j}$
ha nemmeno detto ``ciao''.\\
\> \> \> {\bf He}$_i$ not {\bf to-him}$_j$ has even said ``hi''.\\
\> D2) \> a) Ieri  {\bf Carlo}$_i$ ha incontrato {\bf Maria}$_j$. \\
\> \> \> Yesterday {\bf Carlo}$_i$ has met {\bf Maria}$_j$. \\
\> \> b) $\phi_{*i/j}$  Non {\bf gli}$_{i/*j}$
ha nemmeno detto ``ciao''.\\ \label{exa:carlo2}
\> \> \> {\bf She}$_j$ not {\bf to-him$_i$} has even said ``hi''.
\end{tabbing}
\end{exa}
\unboldmath
\normalsize
Calabrese's analysis correctly explains the allowed and disallowed
coreferences in D1: {\sf Mario} is not the Thema of D1.a. So, if
we want to have the subject of D1.b refer to {\sf Mario}, we cannot use a 
weak pronoun, but we  have to
use a strong one:
in fact, if we do use 
a null subject, it is interpreted as referring to {\sf Carlo}.

Let's now consider D2. The structure of the two discourses is exactly the
same. Therefore the theory predicts that, if we want to refer
to {\sf Maria}, which is not the Thema of D2.a, we have to use a strong 
pronoun, and not a null one: instead, 
D2.b is almost perfect.\\
The reason is that in D2.b the null
subject has two potential referents, one male and the other female.
While processing the sentence, the possibility that the null
subject refers to {\sf Carlo}  is ruled out
when the clitic {\sf gli}, marked for masculine, is found.
In fact, {\sf gli} has to refer to {\sf Carlo}; 
given that {\sf gli} is not reflexive, it cannot corefer with the subject,
therefore the latter is forced to refer to {\sf Maria}.\\
This kind of disambiguation cannot be performed 
in D1.b, in which the null subject has two potential referents of the same
gender.

I should mention that Calabrese, at the beginning of his paper, says that
{\em such features [gender, number and person] allow a first selection
among the possible referents which are assigned to the pronominal}. 
Presumably he would use these features as a superimposed filter
to be applied to the whole sentence after it has been completely 
read or heard.

However, this could hardly fit in a model of how people process
discourse: it is very likely that the normal human mode of operation
is incremental \cite{proc}.  My claim is that disambiguation clues
have to be taken into account as soon as they are available while
processing a sentence. We will see in fact that they can help to make
a discourse coherent or not according to their position in the
sentence.

Notice that the issue here is to account not so much for the
grammaticality or ungrammaticality of a sentence, as pure syntactic
accounts do, but for more or less coherence in a discourse:
this is exactly the purpose of centering theory.
In particular, centering  relates discourse coherence with the inference
load that a certain sequence of utterances, and especially a certain
choice of referring expressions, requires on the part of the hearer.

In the next 
section, I'll show how centering theory can be useful to explain 
certain uses of Italian pronouns in discourse, and in turn, how a richer 
pronominal system can help to refine the rules that centering uses.

\section{Centering theory}

It is now widely accepted that discourse is divided into
segments (see for example \cite{webb88}); a discourse is coherent when
its constituent segments exhibit both local coherence - namely,
coherence among the utterances of each individual segment, and global
coherence - namely, coherence among the different segments.

Centering is an account of local coherence: it tries to
determine the entity which an utterance most centrally concerns.
Besides, it assesses the coherence of a discourse in terms of the
different moves that a speaker can do (basically, going on to talk
about the same entity or switching to another one), and in terms of
how these moves are encoded, in particular as far as the choice of
referring expressions is concerned.  According to \cite{gjw86},
discourse coherence is a measure of the inference load a certain
discourse imposes on a hearer. Notice that the view I am taking on
centering is as a theory of discourse production. From \cite{gjw86},
it is not very clear whether centering concerns the production or the
comprehension of discourse.

More technically, there are three moves that a speaker can perform, for every triple
of utterances $U_n, U_{n+1}, U_{n+2}$, belonging to the same segment:\\
\\
\underline{\bf DEF. 1}
\begin{description}
\item[Continuation:] 
$U_n$ and $U_{n+1}$  concern the same entity; it is
likely that $U_{n+2}$ will concern it too.
\item[Retention:] $U_n$ and $U_{n+1}$  concern the same entity, but
it is not likely that $U_{n+2}$ will concern it.
\item[Shifting:] $U_n$ and $U_{n+1}$ concern different entities.
\end{description}

To formalize these concepts, the theory defines as {\sf centers} those
entities that serve to link one utterance to another in the same
segment; an utterance $U_n$ typically has a single {\sf backward
looking center X (Cb)}, and a set of {\sf forward looking centers (Cf's)
$\{Y_1, ..., Y_m\}$}. \\ X, $Y_1$, ..., $Y_m$ are all candidates for
being Cb($U_{n+1})$ (in fact X = $Y_i$, for some {\em i}), and
Cb($U_{n+1}$) will be constrained to belong to the set of Cf's of
$U_n$. Both Cb$(U_n)$ and the set of Cf's$(U_n)$ correspond to
linguistically realized NPs in $U_n$.\\ 
The set of Cf's for a given
utterance $U_n$ is partially ordered; the ordering relation is
affected by syntactic factors. In \cite{gjw86},
the only syntactic element that is
identified in this respect is the subject of $U_n$: it is the most
likely entity to be Cb$(U_{n+1})$, therefore it is the highest ranked
Cf in $U_n$. This assumption is definitely plausible, but it does not
say anything about ordering among the other Cf's. For a more detailed
analysis of the factors affecting Cf's ordering, see Kameyama' s application of centering to Japanese
\cite{kame}, and for  more recent work on this topic, 
\cite{lyn2}. I will not address this problem in the current
paper.

Given that  the Cb corresponds to the entity that an utterance concerns,
the speaker has some choices as far as encoding the Cb goes.
In \cite{gjw86} the following rule {\bf R1} is proposed:\\
\\
{\sf in $U_{n+1}$ the speaker can use }
\begin{itemize}
\item  {\sf a single pronoun, and that is the Cb$(U_{n+1})$;}
\item  {\sf zero or more than one pronoun: then 
Cb$(U_{n+1})$ is }
\begin{itemize}
\item {\sf Cb$(U_n)$ if 
Cb$(U_n)$ is realized in $U_{n+1}$,}
\item {\sf otherwise the highest ranked Cf$(U_n)$ which
is realized in $U_{n+1}$.}
\end{itemize}
\end{itemize}
In order to ensure a coherent discourse, the speaker  has to apply the following rule {\bf R2} as well:\\

{\sf Given Cb$(U_n) =$ X, Cf$(U_n) = \{Y_1>...>Y_m\}$, X $= Y_k$,
for some k, 
1 $\leq$ k $\leq$ m:}\\
{\sf if there are pairs $\{Y_i, Y_j\}$, with $i < j$, s.t. both $Y_i$ and $Y_j$ are
realized in $U_{n +1}$, and if $Y_j$ is realized with a pronoun, then $Y_i$ has to be realized
with a pronoun.}\\

The previous rule requires that a speaker, if s/he chooses to  use a pronoun to refer
to a certain Cf $Y_j$, has to use a pronoun to refer to all the other Cfs realized
in the current utterance and higher in the ordering than $Y_j$.
\begin{flushleft}
This rule accounts for the unacceptability of discourses like (from \cite{gjw86})  \footnote{Notice that the first utterance of a discourse does
not have a Cb.}:
\end{flushleft}
\boldmath
\footnotesize
\begin{exa}
\normalsize
\footnotesize
\begin{tabbing}
\hspace{1cm} 
\= $U_1$) \=  {\bf John}$_i$ wanted to go for a ride yesterday.\\
\> \> {\tt Cf$(U_1)$ = \{John\}}\\
\> $U_2$) {\bf He}$_i$ called up  {\bf Mike}$_j$.\\
\> \> {\tt Cb$(U_2)$ = John} \\
\> \> {\tt Cf$(U_2)$ = \{John $>$ Mike\}}\\
\> $U_3$) {\bf He}$_j$ was annoyed by   {\bf John}$_i$'s call .\\
\label{exa:jm}
\end{tabbing}
\end{exa}
\unboldmath
\normalsize

In $U_3$, {\sf Mike} is referred to with a pronoun; {\sf Mike} was less highly
ranked than {\sf John} as a Cf, therefore, if we want to refer to {\sf John} in
$U_3$, we should also use a pronoun. The fact that in $U_3$ the proper
name {\sf John} is used makes the sequence unacceptable: in fact
substituting {\sf his} to {\sf John's} results in an acceptable
discourse.

After recognizing what Cb$(U_{n+1})$ is,
the hearer
 can derive the kind of 
move that the speaker has performed in the following way:\\
\\
\underline{\bf DEF. 1$'$}
\begin{description}
\item[Continuation:] Cb$(U_{n+1}) = Cb(U_n)$ and Cb$(U_{n+1})$ is the most highly
ranked element in Cf$(U_{n+1})$. 
\item[Retention:] Cb$(U_{n+1}) = Cb(U_n)$ but Cb$(U_{n+1})$ is not the most highly
ranked element in Cf$(U_{n+1})$. 
\item[Shifting:] Cb$(U_{n+1}) \neq Cb(U_n)$ \footnote{Other versions of centering
provide for two different types of shifting \cite{lyn1}.}.
\end{description}

Notice the correspondence between Def. 1 and Def. 1$'$: the notion of $U_n$ and $U_{n+1}$
concerning the same entity corresponds to $U_n$ and $U_{n+1}$ having the same Cb.
The notion of $U_{n+2}$ going on to concern still the same entity
corresponds to Cb$(U_{n+1})$ being the most highly ranked  Cf$(U_{n+1})$.

\section{Centering and Italian pronouns}

I now want to recast  the choices that the two
Italian pronominal systems offer to a speaker  in terms of centering,
and, at the same time,
refine centering itself.
I will get evidence from examples
like the following \footnote{I am using  referents of different gender, because
I want to show how gender and morphological markings come into play when 
resolving reference. Notice that these examples would not be ambiguous in
English, given that null subject is not an option available to a speaker:
the subject {\sf he/she} would unambiguously pick up its 
referent.}:

\boldmath
\footnotesize
\begin{exa}
\normalsize
\footnotesize
\begin{tabbing}
\hspace{1cm} 
\= $U_1$) \=  {\bf Maria}$_i$ voleva andare al mare.\\
\> \>  {\bf Maria}$_i$ wanted to go to the seaside.\\
\> $U_2$) $\phi_i$ Telefono' a  {\bf Giovanni}$_j$.\\
\> \> {\bf She}$_i$ called {\bf Giovanni}$_j$ up.\\
\> $U_3$) \= a) \= $\phi_i$ Si arrabbio' perche' $\phi_i$ non {\sf lo}$_j$ \\
\> \> \>  trovo' a casa.\\
\> \> \> {\bf She}$_i$ got angry because {\bf she}$_i$ not  {\bf him}$_j$ \\
\> \> \> found at home.\\
\> \> b) $\phi_{i/?j}$ Si arrabbio' \\
\> \> \> perche'  $\phi_j$ stava dormendo.\\  \label{exa:things}
\> \> \> {\bf She}$_i$/ {\bf ?He}$_j$ got angry \\
\> \> \> because {\bf he}$_j$ was sleeping.\\
\> \> c) {\bf Lui}$_j$ si arrabbio' perche' $ \phi_j$ stava dormendo.\\
\> \> \> {\bf He}$_j$  got angry because {\bf he}$_j$ was sleeping.\\
\> \> d) $\phi_j$ Si e' arrabbiat{\bf O}  \\
\> \> \> perche'  $\phi_j$ stava dormendo.\\
\> \> \> {\bf He}$_j$ has gotten angry(-masc.)\\
\> \> \>  because {\bf he}$_j$ was sleeping.\\ 
\end{tabbing}
\end{exa}
\unboldmath
\normalsize

Various interesting facts come out from the four $U_3$ variations \footnote{As
a warning to the reader, notice that I am not worrying about the
interpretation of the null subject in the subordinate causal clause,
as it does not affect the interpretation of the null subject in the
main clause, and it is affected by pragmatic reasons.}:
\begin{description}
\item[{[a]}] The null subject refers to {\sf Maria},
who, according to the rules in the previous section, is Cb$(U_3.a)$,
and the highest ranked element in Cf$(U_3.a)$. $U_3$.a thus
demonstrates center continuation. The discourse is perfectly coherent.
\item[{[b]}] The most natural interpretation is that the null subject
in the main clause refers to {\sf Maria} - the null subject in the subordinate 
clause
is forced to refer to {\sf Giovanni} on pragmatic grounds. 

However, for this same pragmatic reason, on second thought
the null subject in the main clause may be interpreted as referring
to {\sf Giovanni}, but the discourse sounds less coherent. 
\item[{[c]}] The speaker performs  a felicitous center shifting by referring 
 to {\sf Giovanni} with an overt pronoun, given that
{\sf Giovanni} was not Cb$(U_2)$, and not even the highest Cf$(U_2)$.

\item[{[d]}] Contrast this utterance with [b].
They should have the same effect on the hearer,
namely, the null subject should be interpreted as referring to {\sf Maria}:
instead in [d] it is felicitously interpreted as referring to {\sf Giovanni}. 
This happens because
in [d] the verb is in the present perfect tense  \footnote{The temporal
relation between the preceding discourse and [d] is not right; 
$U_2$ should also be in the past perfect. However, this temporal
incoherence 
does not affect resolution of pronoun reference.}; the past
participle agrees with the subject, and its masculine morphology
forces the referent of the null subject to be {\sf Giovanni}, and not
{\sf Maria}.

\end{description}

It seems to me that Ex.\ref{exa:things} and other similar examples point to the following
generalizations:
\begin{itemize}
\item typically, the speaker encodes  center continuation with a null subject.
This agrees with Kameyama's analysis of Japanese \cite{kame});
\item he typically encodes center retention
or shift with a stressed pronoun;
\item he can felicitously use a null subject  in cases of center retention or shift 
if  he provides $U_{n+1}$ with syntactic features that force the null subject to 
refer to a particular referent and not to Cb$(U_n)$. 
\end{itemize}

My claim is that it is the syntactic context up to and including the
verbal form(s) carrying tense and / or agreement that makes the
reference felicitous or not.  Consider $U_3$.d again: it is the fact
that the main verb is marked for masculine that allows the null
subject to refer to something different from Cb$(U_2)$.

Analogous considerations hold for D2.b in Ex. \ref{exa:carlo2}. There the clitic {\sf 
gli} precedes the verb and forces the null subject to refer to {\sf Maria}. 
The fact that the clitic precedes the verb is crucial: evidence for this derives from examples involving modal verbs and clitics.

\footnotesize
\begin{exa}
\normalsize
\footnotesize
\begin{tabbing} 
\hspace{1cm}
\= $U_1$) \= {\bf Maria}$_i$ e' arrabbiata con  {\bf Giorgio}$_j$:\\
\> \> {\bf Maria}$_i$ is angry with {\bf Giorgio}$_j$:\\
\> $U_2$) \= a) \= $\phi_i$ non vuole piu' parlar{\bf gli}$_j$.\\
\> \> \>  {\bf she}$_i$ not wants any more talk-{\bf to-him}$_j$.\\
\> \> b)  * $\phi_j$ non vuole piu' parlar{\bf le}$_i$.\\
\> \> \>  * {\bf he}$_j$ not wants any more talk-{\bf to-her}$_i$.\\
\> \> c)  $\phi_j$ non {\bf le}$_i$ vuole piu' parlare.\\
\> \> \>  {\bf he}$_j$ not {\bf to-her}$_i$ wants any more talk. 
\end{tabbing}
\end{exa}
\normalsize

Here  $U_2$.a is perfect, with the null subject referring to the higher Cf$(U_1)$, namely {\sf Maria}. \\
$U_2$.b is incoherent: the null subject is interpreted as referring to {\sf Maria},
but when the clitic {\sf le} is found, at the end of the sentence, the 
hearer is forced to change interpretation. The effect is similar to a 
syntactic ``garden path''.\\  
$U_2$.c is acceptable, for the very reason that the clitic {\sf le},
that in $U_2$.b is cliticized onto {\sf parlare},
climbs in front of the modal verb {\sf vuole}: so 
the hearer
is forced to exclude {\sf Maria} as referent of  the null subject. This happens  {\em early enough}
so that no ``garden path'' effect is registered.

\section{Other phenomena}

The predictions presented in the previous section 
are quite reliable, but there 
are some cases that are not taken into account.

\subsection{Purpose of an utterance}
Consider the following example:
\boldmath
\footnotesize
\begin{exa}
\normalsize
\footnotesize
\begin{tabbing}
\hspace{1cm} 
\= $U_1$) \= {\bf Luisa}$_i$ ha lasciato {\bf suo marito}$_j$: \\
\> \> {\bf Luisa}$_i$ has left {\bf her husband}$_j$:\\
\> $U_2$) $\phi_{*i/j}$  picchiava i bambini e si ubriacava.\\ \label{exa:luisa}
\> \> {\bf he}$_j$ used to beat the children\\
\> \>  and get drunk.
\end{tabbing}
\end{exa}
\unboldmath
\normalsize
In this case, $Cb(U_2)$ is Luisa's husband. $U_2$ is felicitous, although 
the speaker uses a null subject to achieve a shift and  no syntactic clue
 forces the null subject not to refer to Luisa. It looks like it is the function of
$U_2$, namely, explaining why Luisa  left her husband, that licenses the use
of a null subject in this case.

It may even be argued that this case is outside the purview of
centering, which explicitly states that the referential phenomena
accounted for are within a single segment: $U_2$ may belong to a new
segment, possibly much longer than what is shown here, that explains
why Luisa left her husband.

On the other hand, it seems to me that
the concept of local coherence is not totally dependent on having two
utterances belonging to the same segment. The transition to another
segment may override centering predictions; nevertheless, the
referential expressions found in the first utterance of the new
segment may need to be accounted for in terms of the Cf's of the last
utterance of the previous segment.  This may be what happens in
Ex.\ref{exa:luisa}, if indeed $U_2$ belongs to a new segment.

\subsection{Null subject referring to a whole discourse segment}

Reference to a whole discourse segment is generally achieved in
Italian by means of {\sf questo / cio'}, both equivalent to {\sf
this}, but sometimes a null subject is used (on this topic, see
\cite{pennling89}):

\boldmath
\footnotesize
\begin{exa}
\normalsize
\footnotesize
\begin{tabbing}
\hspace{1cm} \= Questi grandi atleti sono illuminati dai \\
\> {\sl These great athletes come under} \\
\> mass media ogni due, ogni quattro anni,\\
\> {\sl the media light every two, every four years,}\\
\> e devono conquistare una medaglia  \\
\> {\sl and they have to win a medal} \\
\> lottando contro il mondo intero  \\
\> {\sl fighting against the whole world}\\
\> per guadagnarsi l'affetto della gente.  \\
\> {\sl to gain people's affection.} \\
\> Mentre in altri sport (nel calcio soprattutto) \\
\> {\sl In other sports (in soccer above all),}\\
\> l'amore, la celebrita', i denari \\
\> {\sl love, fame, money} \\
\> sono quasi automatici, quasi obbligatori.  \\
\> {\sl are almost automatic, almost compulsory.} \\
\> $\phi$ E' giusto? \label{exa:exsport}\\
\> {\sl Is {\bf this} fair?}
\end{tabbing}
\end{exa}
\unboldmath
\normalsize

In the preceding example, the null subject in the last utterance
refers to the whole previous discourse: the fact that a null subject,
namely, the pronoun with the least informative content,  that should
supposedly
refer to an expected referent, can be used in
such a way, is a phenomenon that deserves explanation.

In general, centering does not say anything about reference to
discourse segments, and in fact it may again be argued that clausal
reference has nothing to do with local coherence.\\ This actually
depends on the perspective from which we look at clausal reference: it
is possible that entities corresponding to discourse segments are
implicitly included in the Cf's set; or that they are available for
reference, but they have a status different from the normal Cf's; or
that they have a different status altogether, for example that they
do not exist as centered entities until they are referred to for the
first time \cite{webb88}.

In any of these three cases, a theory of discourse coherence should at
least partly address the problem.

\section{Conclusions}

In this paper, I have shown how the context up to and including the verb
helps in disambiguating the reference for a null subject.

Some topics for future research have been discussed in the previous 
section. Integrating the analysis of these phenomena with centering will 
shed some light on the whole phenomenon of reference. 

Centering gives us a vintage point of view in looking at local
coherence in discourse as embodied by the choice of referring
expressions that a speaker uses.  Languages with
richer morphological marking and agreement system than English 
can be very useful both to assess centering and to refine its rules.

\begin{flushleft}
{\bf Acknowledgements.} 
\end{flushleft}
I would like to thank Prof. Bonnie Webber for her support
and her comments on earlier versions of this paper,
and Prof. Aravind Joshi for useful discussions.

\end{document}